\def\year{2021}\relax
\documentclass[letterpaper]{article} 
\usepackage{aaai21}  
\usepackage{times}  
\usepackage{helvet} 
\usepackage{courier}  
\usepackage[hyphens]{url}  
\usepackage{graphicx} 
\def\UrlFont{\rm}  
\usepackage{natbib}  
\usepackage{caption} 
\usepackage[utf8]{inputenc} 
\usepackage[T1]{fontenc}    
\usepackage{booktabs}       
\usepackage{amsfonts}       
\usepackage{nicefrac}       
\usepackage{microtype}      

\usepackage{tabularx}
\usepackage{graphicx}
\usepackage{amsmath} 
\usepackage[utf8]{inputenc}
\usepackage{textcomp}
\usepackage{url}
\def\UrlBreaks{\do\/\do-}
\usepackage[hang,flushmargin]{footmisc}
\usepackage{color}
\usepackage{xcolor}
\usepackage{makecell}
\usepackage{diagbox}
\usepackage{enumitem}
\usepackage{multirow}

\usepackage{pgfplots}
\usepgfplotslibrary{colorbrewer}
\usepgfplotslibrary{dateplot}
\usepgfplotslibrary{fillbetween}
\pgfplotsset{compat=1.14}
\pgfplotsset{cycle list/Dark2}

\newcommand{\minisection}[1]{\vspace{0.05in} \noindent {\bf #1}}

\title{
  Deep policy networks for NPC behaviors that adapt to changing design parameters in Roguelike games
}

\author{
    Alessandro Sestini,\textsuperscript{\rm 1} 
    Alexander Kuhnle,\textsuperscript{\rm 2} 
    Andrew D. Bagdanov\textsuperscript{\rm 1}
    \\
}
\affiliations{
    \textsuperscript{\rm 1}Dipartimento   di   Ingegneria   dell’Informazione,    Università degli  Studi  di  Firenze, Florence,  Italy\\
    \textsuperscript{\rm 2}Department  of  Computer   Science   and   Technology,   University of   Cambridge, United  Kingdom\\ 
    \{alessandro.sestini, andrew.bagdanov\}@unifi.it, alexander.kuhnle@cantab.net
}

\begin{document}

\maketitle

\begin{abstract}
  Recent advances in Deep Reinforcement Learning (DRL) have largely
  focused on improving the performance of agents with the aim of
  replacing humans in known and well-defined environments. The use of
  these techniques as a game design tool for video game production,
  where the aim is instead to create Non-Player Character (NPC)
  behaviors, has received relatively little attention until
  recently. Turn-based strategy games like Roguelikes, for example,
  present unique challenges to DRL.  In particular, the categorical
  nature of their complex game state, composed of many entities with
  different attributes, requires agents able to learn how to compare
  and prioritize these entities. Moreover, this complexity often leads
  to agents that overfit to states seen during training and that are
  unable to generalize in the face of design changes made during
  development. In this paper we propose two network architectures
  which, when combined with a \emph{procedural loot generation}
  system, are able to better handle complex categorical state spaces
  and to mitigate the need for retraining forced by design
  decisions. The first is based on a dense embedding of the
  categorical input space that abstracts the discrete observation
  model and renders trained agents more able to generalize. The second
  proposed architecture is more general and is based on a Transformer
  network able to reason relationally about input and input
  attributes. Our experimental evaluation demonstrates that new agents
  have better adaptation capacity with respect to a baseline
  architecture, making this framework more robust to dynamic gameplay
  changes during development. Based on the results shown in this
  paper, we believe that these solutions represent a step forward
  towards making DRL more accessible to the gaming industry.
\end{abstract}


\section{Introduction}
\label{sec:intro}
\begin{figure}
    \centering \includegraphics[width=0.8\columnwidth]{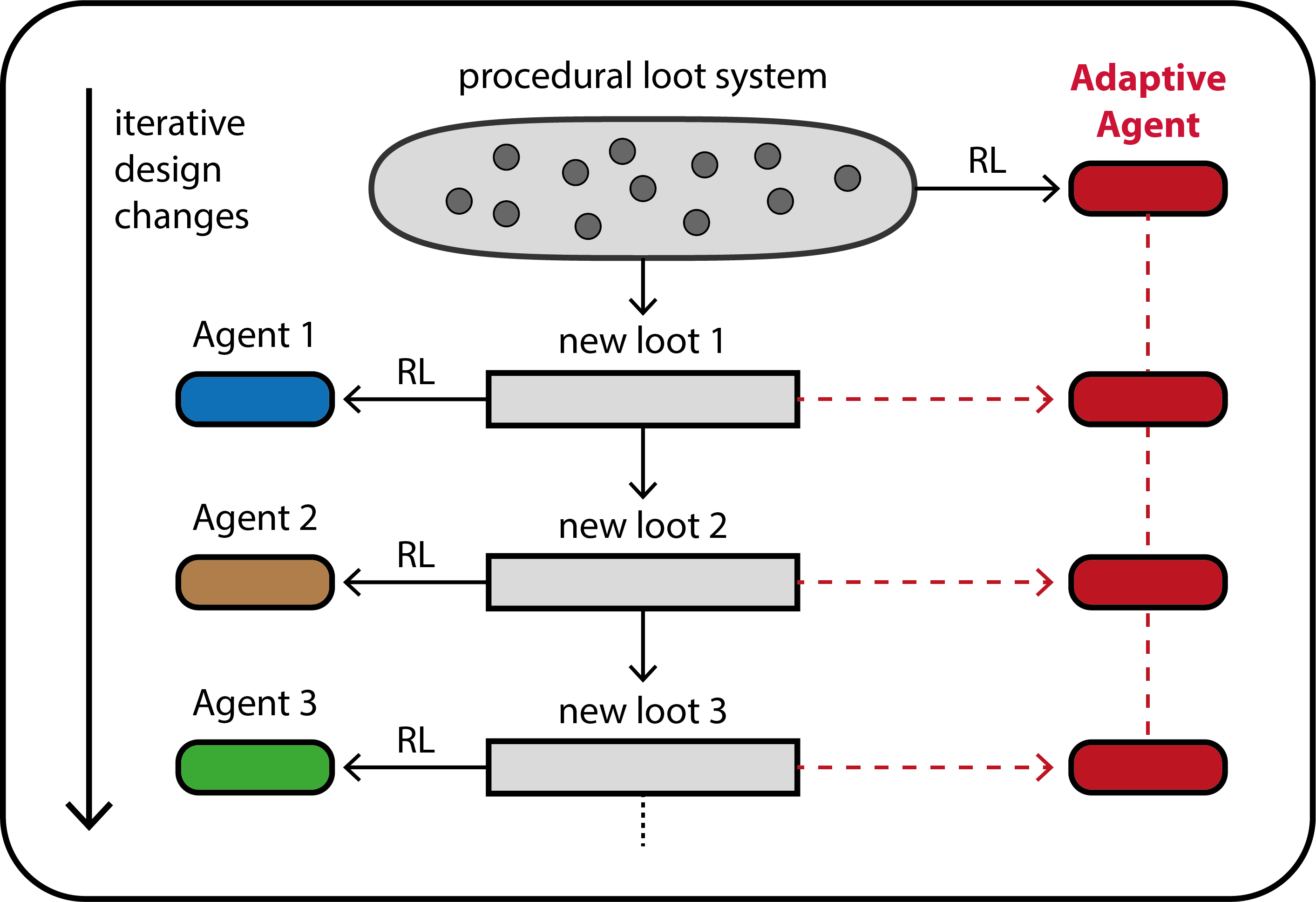}
    \caption{A summary of our approach. The left side illustrates the original DeepCrawl
    framework in which the agent must be retrained every time the
    loot distribution changes (e.g. for balancing the
    overall game). Our approach is shown on the right: 
    with the new adaptive architectures we can create agents which can 
    learn from a new procedural loot system. Agents trained in this way 
    are able to adapt to a changing loot distribution without the need 
    to retrain.
    }
    \label{fig:teasing}
\end{figure}

In the gaming industry, Artificial Intelligence (AI) systems that
control Non-Player Characters (NPCs) represent a vital component in
the quality of games, with the potential to elevate or break the
player experience. Recently, examples of NPC agents trained with Deep Reinforcement Learning (DRL) techniques have been
demonstrated for commercial video games, however mass adoption by game
designers requires significant technical innovation to build trust in
these approaches \cite{microsoft}.  A step in this direction was the
DeepCrawl prototype \cite{sesto19}, a Roguelike game where all NPCs
were moved by DRL algorithms.

In this paper we address the challenges of \textit{adaptation} and
\textit{scaling}, described by the aforementioned authors, and also encountered
in DeepCrawl. DRL algorithms are extremely sensitive to design changes
in the environment, since they fundamentally change the way agents
``see'' the game world around them. Even seemingly minor changes can
force a complete retraining of all agents. This is mostly due to the
categorical nature of the input state space which makes the network
overfit to the specific entities seen during training, leaving it
without the capacity to generalize to unseen states.  Collectible
objects in DeepCrawl and their effect on the game, for example, must
be predefined by developers, and are represented by unique integer IDs
and not by their effect on the player. This can be an important
problem during game development: if developers want to change
parameters, for example to balance gameplay, they require agents which
can handle these modifications and do not require retraining. This
makes it difficult to adapt an existing agent to new scenarios,
resulting in inappropriate agent behavior when NPC agents are used in
environments for which they were not designed.

Moreover, with the NPC model architecture of the original DeepCrawl work 
it is not possible to extend the set of available loot or loot types without
completely retraining agents from scratch. This is largely due to specific
DeepCrawl network architecture: the policy network contains initial
\textit{embedding layers} that make it possible for the network to learn a
continuous vectorial representation encoding the meaning of and differences
between \emph{categorical} inputs. As mentioned above, in this setting each loot
item must be identified by a unique ID in order to be understandable by agents.
For this reason, if designers want to add new loot types, for example changing
the object definition in order to have a different number of attribute bonuses,
it is difficult or impossible to define a unique ID for each object \emph{a
  priori} -- particularly if the attribute bonuses are determined randomly
during game play.

To mitigate these problems we implemented a new \textit{procedural loot
  generation} system and incorporated it into the training protocol:
instead of a fixed list of discrete items, in our new system an item is
parametrized by a fixed set of attributes, potentially even an extensible set of
attributes. These values are drawn from a uniform distribution when generating a
new training episode and increase the intrinsic properties of actors when
collected. We also propose two alternative policy network architectures that are
able to handle the new procedural loot system. As we illustrate in figure
\ref{fig:teasing}, the new system combined with procedural loot generation
during training renders trained NPCs more adaptive and scalable from a
game developer's perspective. This new AI system helps in the design
of NPC agents while being robust to iterative design changes across
the loot distribution that can happen during video game
development. As our experiments show, these new agents are
perfectly capable of adapting to loot distributions they have never seen
during training, without the need to retrain.


\begin{figure}
  \begin{center}
    \centerline{\includegraphics[width=0.65\columnwidth]{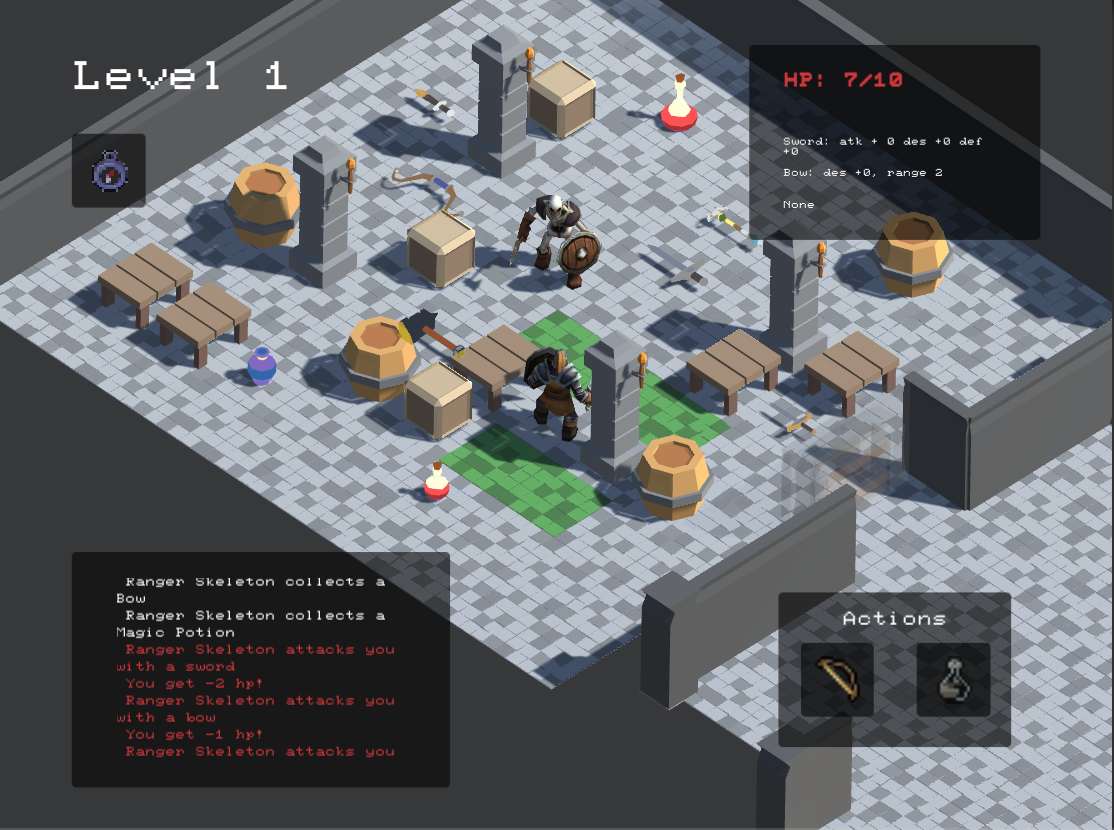}}
    \caption{Screenshot of the DeepCrawl game. For more information
    about gameplay elements see section \ref{sec:deepcrawlnet} and the original DeepCrawl paper \cite{sesto19}.}
    \label{fig:screenshot}
  \end{center}
\end{figure}

\section{Related work}
\label{sec:related}
The potential of DRL in video games has been steadily gaining interest
from the research community. Here we review recent works most related
to our contributions.

\minisection{Procedurally Generated Environments.}  There is a growing
interest in DRL algorithms applied in environments with Procedural
Content Generation (PCG) systems: \cite{procenv} demonstrated that
diverse environment distributions are essential to adequately train
and evaluate RL agents, as they observed that agents can overfit to
exceptionally large training sets.  On the same page are
\cite{procedural}, who stated that often an algorithm will not learn a
general policy, but instead a policy that only works for a particular
version of a particular task with specific initial parameters.
\cite{illuminating} explored how procedurally generated levels during
training can increase generalization, showing that for some games
procedural level generation enables generalization to new levels
within the same distribution. Subsequently, the growing need for a PCG
environments was also demonstrated by \cite{nethack}, \cite{minigrid},
and \cite{unitytower}.

\minisection{DRL \emph{in} video games.} Modern video games are
environments with complex dynamics, and these environments are useful
testbeds for testing complex DRL algorithms. Some notable examples
are: \cite{alphastar} that uses a specific deep neural network
architecture based on Transformers \cite{transformers} able to create
super-human agents for StarCraft, and \cite{openai2019dota} that use
embedding layers similar to \cite{sesto19} to manage the inner
attributes of the agent and other heroes in DOTA 2 in order to train
agents that outperform human players.

\minisection{DRL \emph{for} video games.} At the same time, there is
an increasing interest from the game development community on how to
properly use DRL for video game development.  \cite{elettronicarts}
argued that the industry does not need agents built to ``beat the
game'', but rather to produce human-like behavior to help with game
evaluation and balance. \cite{ubisoft} dealt with the importance of
having an easy-to-train neural network and how it is important to have
a framework that enriches the expressiveness of the
policy. \cite{unityaction} studied different action-space
representations in order to create agents that mimic human input,
without being super-human.  As already discussed, \cite{sesto19}
contributed to this aim, defining a DRL framework suited for the
production of turn-based strategy games. Our aim is to improve on the
latter framework in order to render it more robust to changes to
gameplay mechanics during development -- i.e., to render DRL agents
more \emph{mechanics-free}.


\section{Proposed models}
\label{sec:models}
\label{sec:desiderata}

Our work builds upon the DeepCrawl framework. Our overarching goal is
to make the system as independent as possible from dynamic changes
during the development phase, and we argue that a crucial step in this
direction is a \textit{procedural loot generation system} which helps
encourage generalizing agent behavior in a fully procedural
environment. In particular, we want to fulfill the following
desiderata:
\begin{itemize}
\item \textbf{Performance.} We desire agents able to properly handle a
procedural loot system, so they must understand which object is most
useful for defeating the game;

\item \textbf{Adaptation.} Agents must adapt to changes in gameplay
  mechanics, in particular changes to the loot generation system
  during playtesting and rebalancing, without the need of retraining;
  and

\item \textbf{Scalability.} We desire a framework that can scale in both the
  number of possible objects and in the number of attribute bonuses of
  each object type. Moreover, the framework must have limited
  complexity to facilitating targeting of systems like mobile devices.
\end{itemize}

With these three new desiderata in mind, we now describe two
architectural solutions that satisfy them. Both are significant
modifications of the early, frontend layers of the DeepCrawl network
that allow it to better manage our new procedural loot system. We
begin with a brief introduction of the original DeepCrawl environment
and network, and then continue with the description of two different
architectures that address the problems defined above.

\begin{figure*}
  \vskip
  0.2in \begin{center} \centerline{\includegraphics[width=1.50\columnwidth]{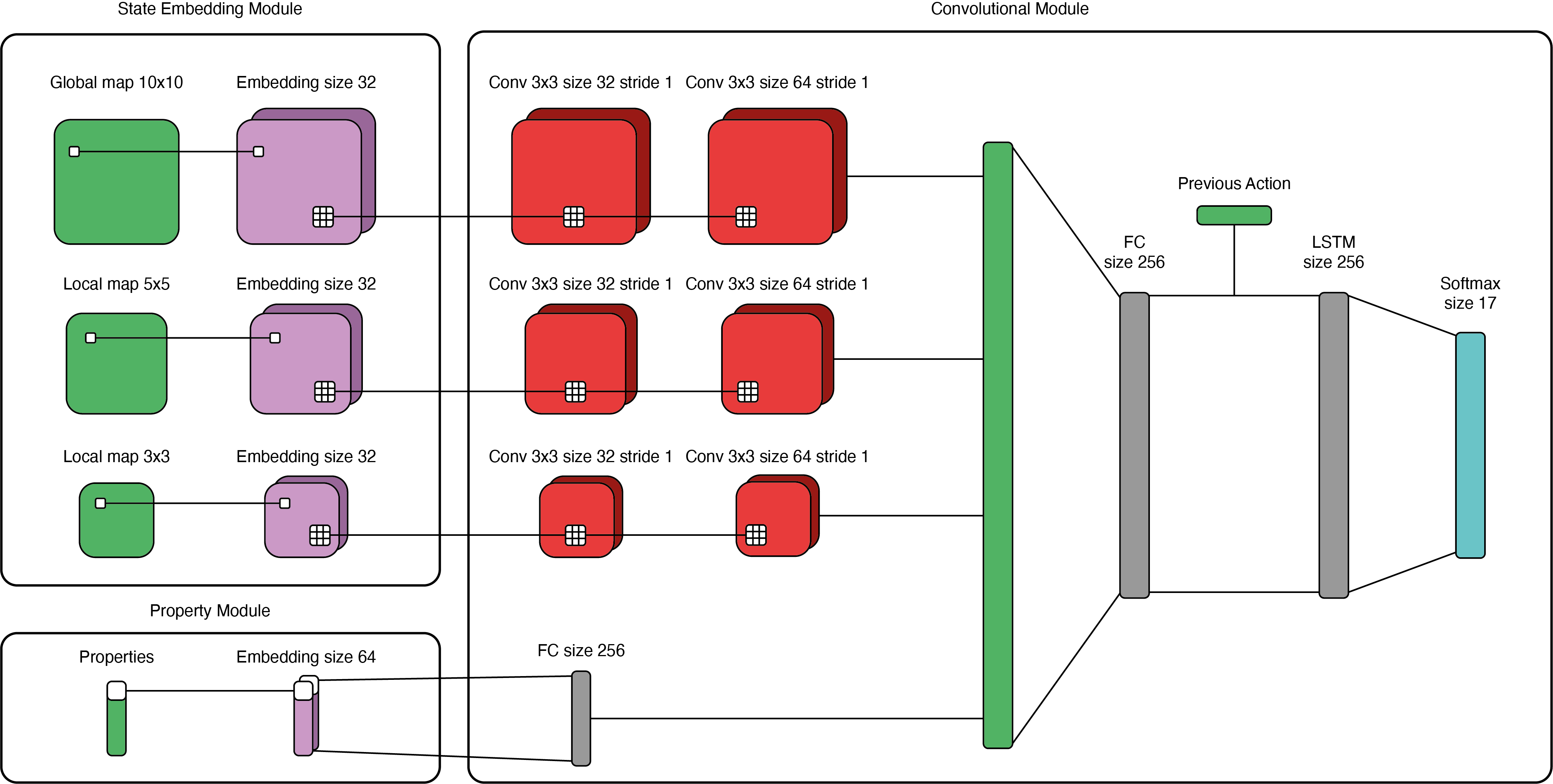}} \caption{ 
  The Categorical network used for NPCs in DeepCrawl (see
  section~\ref{sec:deepcrawlnet} for a detailed description). This
  network architecture is not able to properly handle a procedural
  loot generation system due to the \textit{State Embedding Module}
  that requires categorical IDs for each entity in the
  game. Our approach replaces this module with two possible
  alternatives shown in figure \ref{newnets}.
  } \label{deepcrawlnet} \end{center} \vskip -0.2in
\end{figure*}

\begin{figure*}[ht!]
\begin{center}
\scalebox{1.05}{
\begin{tabular}{cc}
  \includegraphics[width=0.40\textwidth]{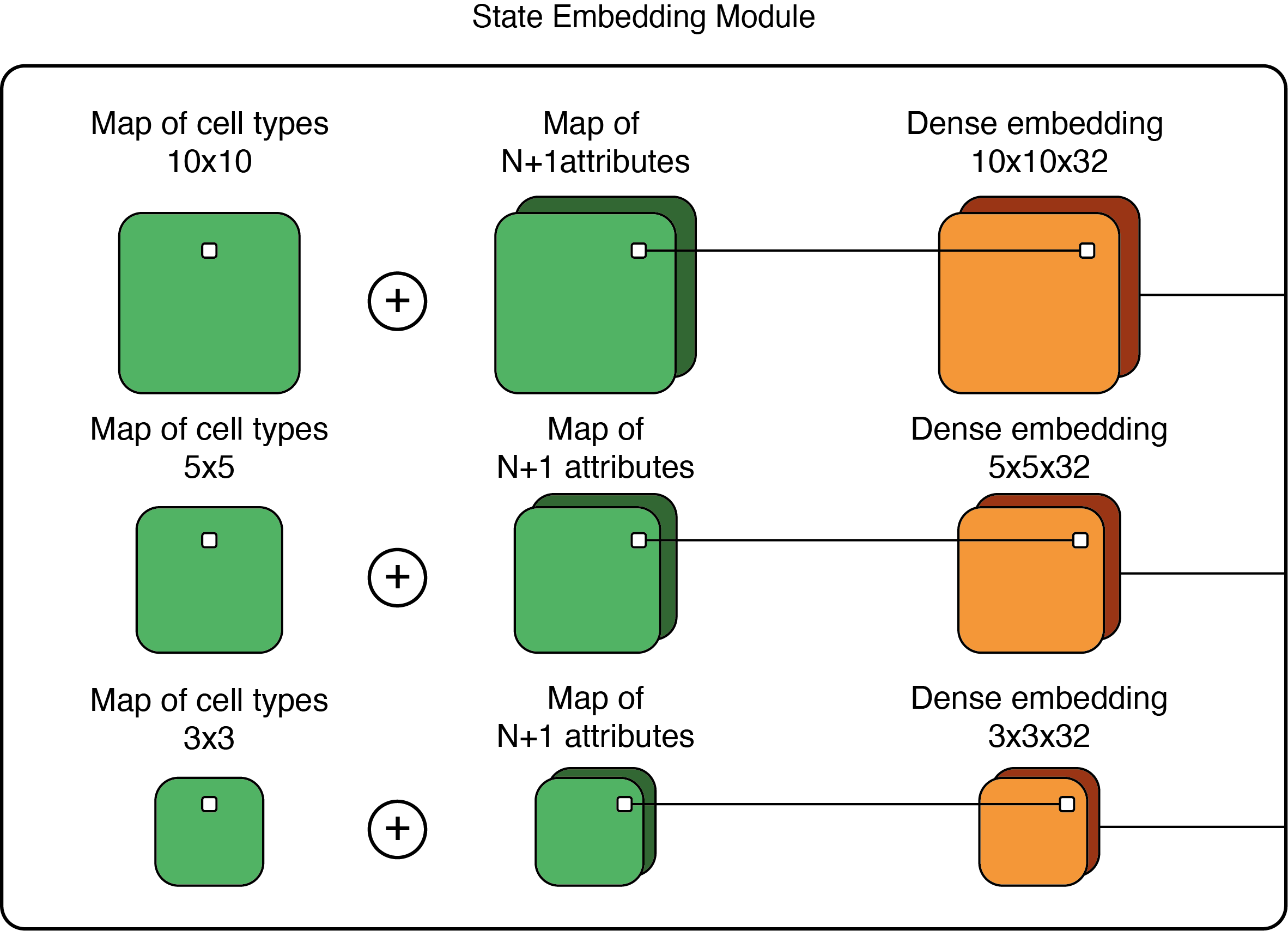} &
  \includegraphics[width=0.40\textwidth]{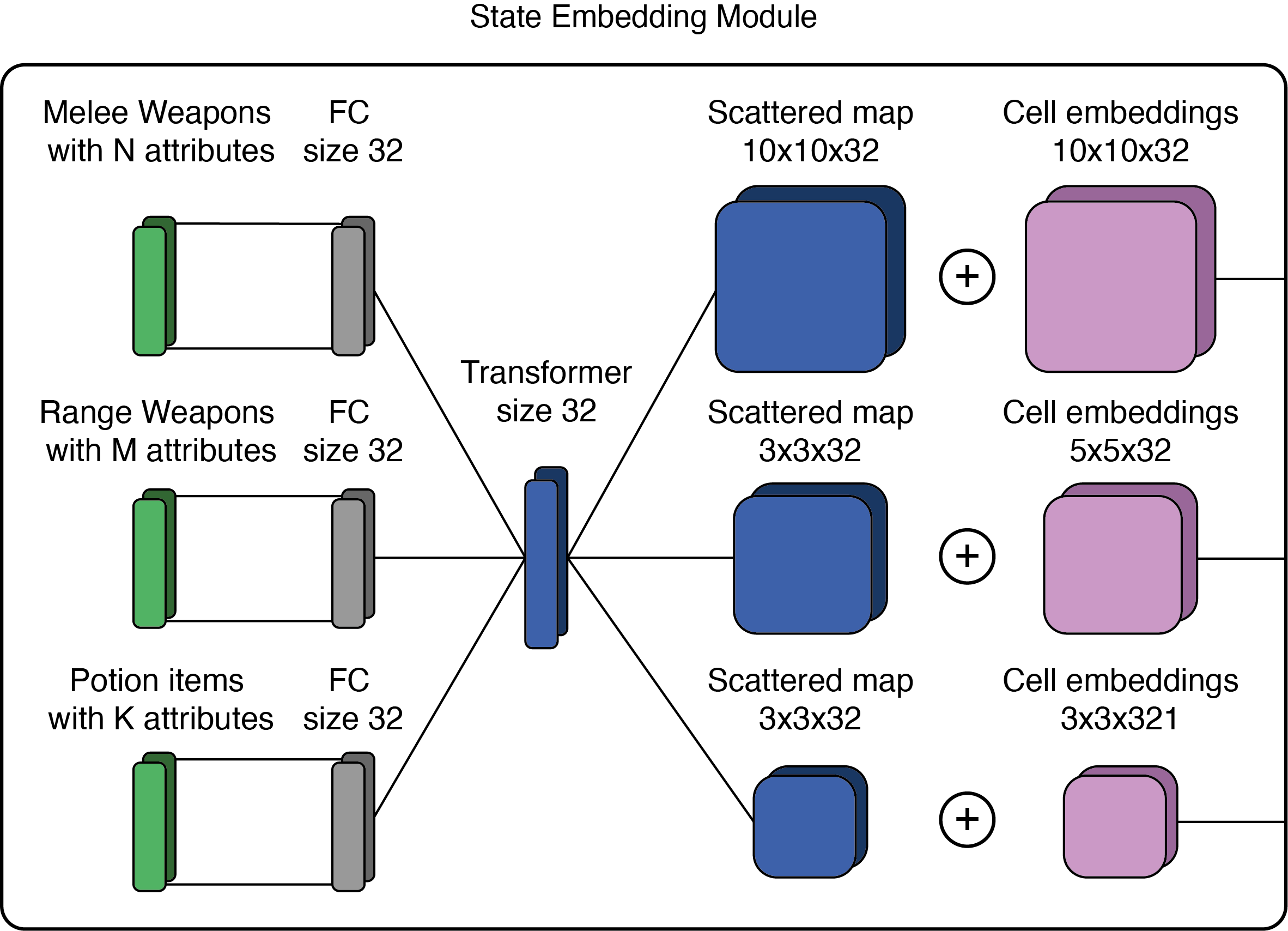} \\
  (a) Dense Embedding module & (b) Transformer Embedding module
\end{tabular}
}
\end{center}
\caption{The two new architectures proposed in this paper: (a) The
  Dense Embedding module, and (b) the Transformer Embedding
  module. These modules replace the \textit{State Embedding Module} in
  the original architecture shown in figure \ref{deepcrawlnet}, while
  the rest of the policy network is left unchanged. See
  section~\ref{sec:densenet} for a detailed description.}
\label{newnets}
\end{figure*}

\subsection{The DeepCrawl environment and policy network}
\label{sec:deepcrawlnet}
DeepCrawl (figure \ref{fig:screenshot}) is a \emph{Roguelike} game
that shares all the typical elements of the genre, such as
the \textit{procedurally created environment}, the
\textit{turn-based} system, and the \textit{non-modal} characteristic that makes
every action available to actors regardless the level of the game. In
this environment the player faces one or more agents controlled by a
DRL policy network. Player and agents act in procedurally-generated
rooms, and both of them have exactly the same characteristics, can
perform the same 17 actions, and have access to the same information
about the world around them. The reward function is extremely sparse
and only gives positive reward in case of victory. Player and agent
are aware of a fixed number of personal characteristics such as HP,
ATK, DEX, and DEF.

The visible environment at any instant in time is represented by a
grid with maximum size of $10 \times 10$ tiles. Each tile can contain
an agent or player, an impassible object, or collectible loot. Each of
these entities are represent with a categorical integer ID. Loot can
be of three types: melee weapons, ranged weapons, or potions. There is
a fixed set of loot, each of which increases an actor characteristic
by a predefined value according to the type of object. In this
context, the agent must learn which loot ID is the best to have in
order to win the game.

Success and failure in DeepCrawl is based on
direct competition between the player and one or more 
agents guided by a deep policy network trained
using DRL. Player and agents have exactly the same characteristics,
can perform the same actions, and have access to the same information
concerning the world around them.

The input state is divided into one global view, the whole grid map, 
and two local views, smaller maps centered around the agent’s position
at different scales, that are passed as input to a convolutional neural
network. A fourth
input branch takes as input an array of discrete values containing
information about the agent and the player. Due to the categorical
nature of the input state space described above, we call this
architecture a \textit{Categorical network} and the overall details
are illustrated in figure \ref{deepcrawlnet}.
We refer to the first input branches as State Embedding module and 
the fourth input branches as Property module. In this paper we 
focus mainly on the State Embedding module.

As was discussed in the introduction, this input structure limits the
adaptation nature of the trained agents. We overcome this limitation
by first defining and implementing a different way to generate
collectible items in the environment.

\subsection{Procedural loot system}
\label{sec:loot-system}
In our proposed parametric loot system, each object has a number of
attributes whose values during training are drawn from a uniform
distribution when generating an environment for a new training
episode. When an actor collects an item, the actor characteristics
will increase or decrease according to the attributes of the instance
of the looted object. In our current implementation, each object has
the same four attributes corresponding to the four characteristics of
the actors. This is not a requirement, however, rather it reflects the
original design and implementation of categorical loot system in the
original DeepCrawl game.

This system brings a lot of benefits to DeepCrawl: it makes the game
more complex and varied, with the corresponding possibility of
creating more convincing NPCs and player/environment interactions. The
environment is now fully procedural, which should increase the
generalization of the agents. Moreover, during playtesting developers
can choose either to use random objects or to define a set of fixed
objects with fixed attributes in order to balance the game.

However, to enjoy these benefits the policy networks trained for agent
behaviors must be able to accommodate this new procedural loot
system. The network described above cannot easily do this due to the
categorical nature of its input space. Thus we propose two new
solutions.

\subsection{Dense embedding policy network}
\label{sec:densenet}
Our first model is a straightforward extension of the one used in the
original DeepCrawl paper. We were inspired by the ideas
of \cite{openai2019dota} and DeepCrawl to treat the map of categorical
inputs via embedding layers. In contrast to these approaches, however,
we use multi-channel maps where each channel represents a different
categorical value:
\begin{itemize}
\item The first channel represents the type of entity in that position:
  \begin{itemize}
  \item 0 = impassable object;
  \item 1 = empty tile;
  \item 2 = agent;
  \item 3 = player;
  \item 4 = melee weapon;
  \item 5 = range weapon; and
  \item 6 = potion item.
  \end{itemize}
\item The other channels each contain an attribute of the object in
  that tile, represented by a categorical value. For instance, if a
  tile contains a melee weapon, its attribute bonuses like health
  points (HP), attack (ATK), defense (DEF), and dexterity (DEX) are
  represented by an array of all attributes (plus tile
  type): \texttt{[TYPE, HP, ATK, DEF, DEX]}. If the tile does not
  contain loot (like an impassable object), this array is filled with
  the special value \texttt{no-attribute}: \texttt{[TYPE, NONE, NONE,
  NONE, NONE]}.
\end{itemize}

This multi-channel map input, which like the original DeepCrawl
network as shown in figure~\ref{deepcrawlnet}, is divided in global
and local views and passed through what we refer to as \textit{``dense
embedding''} layers: multiple categorical values are combined together
and mapped to their corresponding fixed-size continuous representation
by a single dense embedding operation. To implement the dense
embedding operation, we simply convert each channel into a one-hot
representation and apply a $1 \times 1$ convolution with stride $1$
and $tanh$ activation through all channels. In the special case of a
single channel, the operation is equivalent to standard embedding
layers. The full model architecture is shown in figure \ref{newnets}a.

This architecture satisfies the requirements we are looking for: the
framework is independent from attribute changes to the loot system as
long as the types of character attributes remain the same. Moreover,
if developers want to change the set of attributes during production,
it is no longer necessary to change the entire agent architecture,
only the corresponding channels of the dense embedding layer need to
be added or removed. As an additional benefit, the network size
remains relatively small. A detailed analysis of experimental results
for this architecture are given in section \ref{sec:experiments}.

\begin{figure*}[ht!]
\scalebox{1.0}{
\begin{tabularx}{\textwidth}{XXX}
  \includegraphics[width=0.35\textwidth]{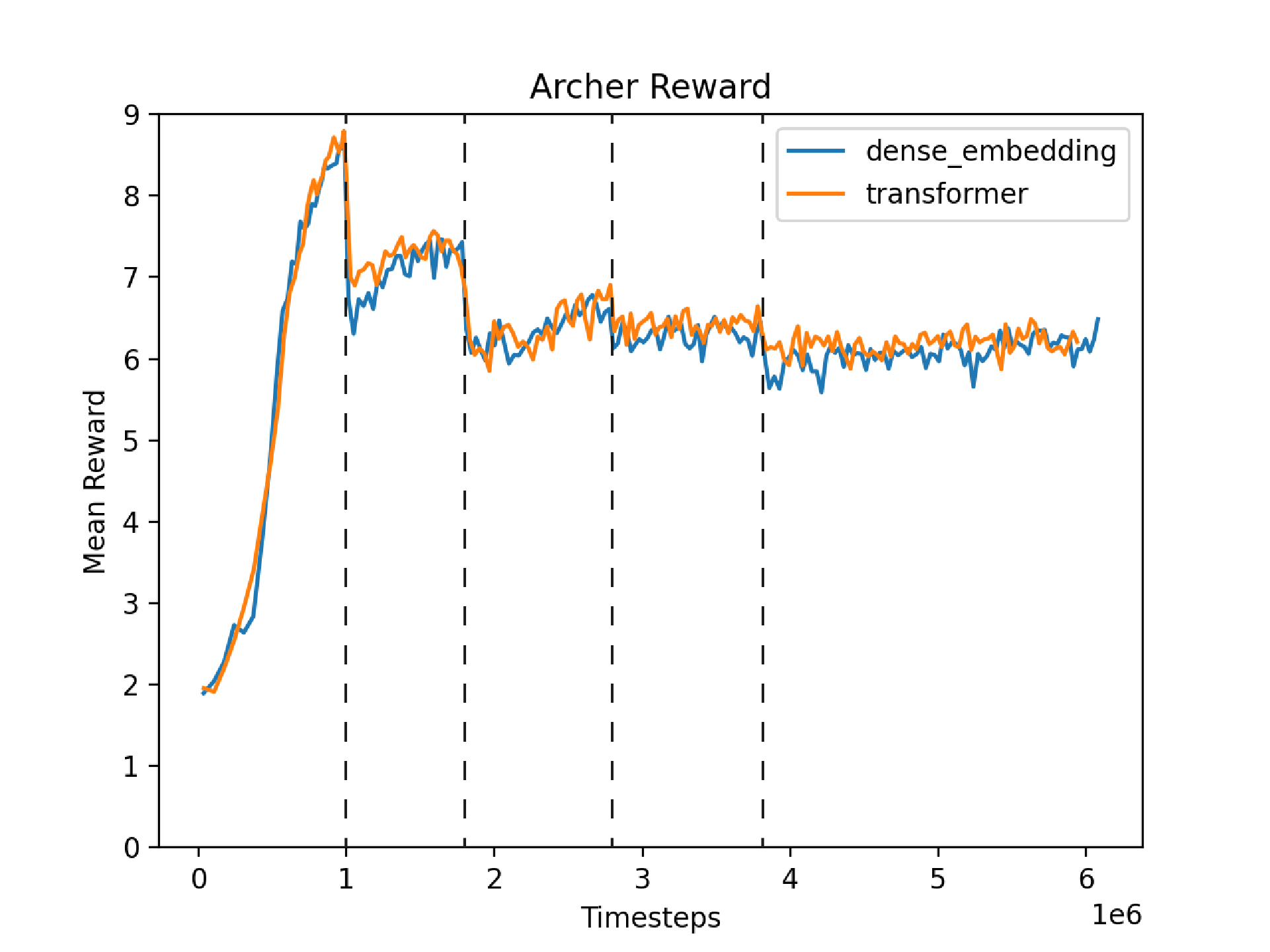} &
  \includegraphics[width=0.35\textwidth]{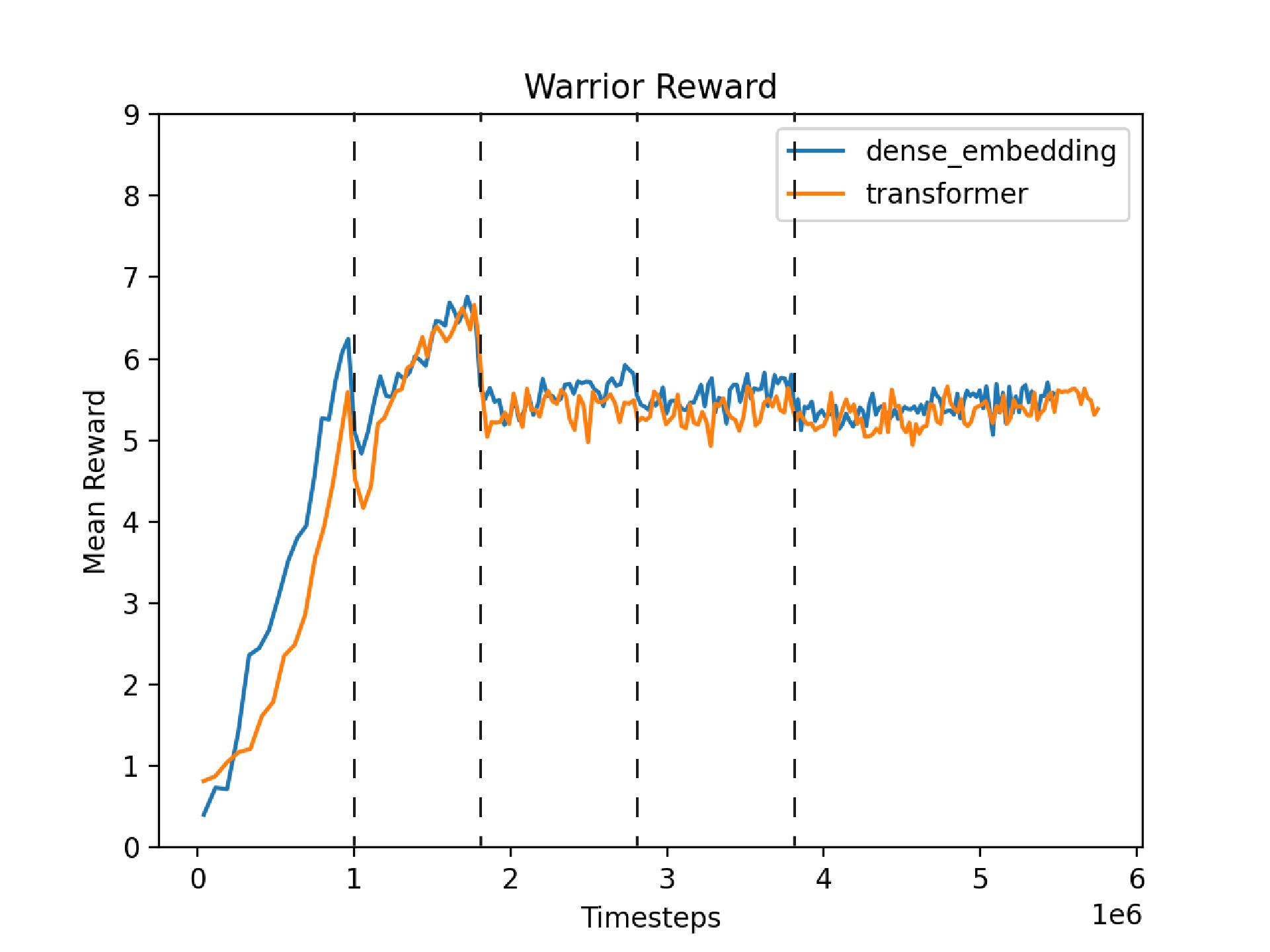} &
  \includegraphics[width=0.35\textwidth]{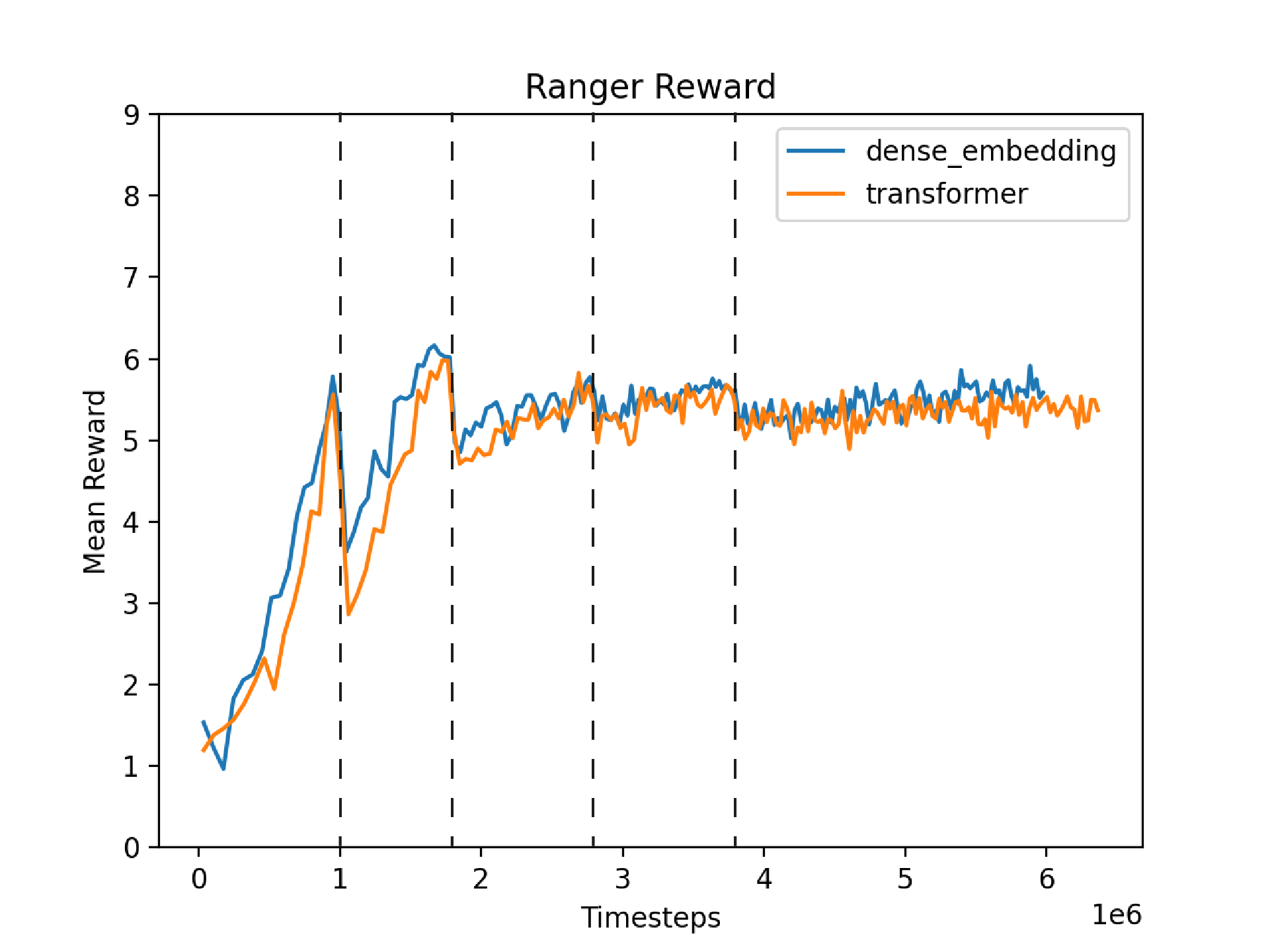} 
\end{tabularx}
}
\caption{ Mean reward during the training phase for all classes as a
  function of timestep. From left to right: archer, warrior, and
  ranger. The dashed vertical lines on the plots delineate the different curriculum
  phases, which are the same as in~\cite{sesto19}.}
\label{training}
\end{figure*}

\subsection{Transformer-based policy network}
\label{sec:transnet}
We propose an alternative model based on the recently popular
  Transformer architecture~\cite{transformers}, and particularly its
  self-attention layer which has also been successfully applied as
  state encoder in RL
  applications \cite{openaitrans,alphastar,relational}.  This model
  uses self-attention to iteratively reason about the relations
  between entities in a scene, and is expected to improve upon the
  efficiency and generalization capacity over convolutions by more
  explicitly focusing on entity-entity relations.

Concretely, the self-attention layer takes as input the set of
entities $e_i$ for which we want to compute interactions (not
including auxiliary \texttt{no-attribute} objects), and then computes
a multi-head dot-product attention
\cite{transformers}: given $N$ entities, each is projected to a query $q_i$, a key $k_i$ and a value $v_i$ embedding, and the self-attention values are computed as
\begin{equation}
A = \mbox{softmax}\left(\frac{QK^t}{\sqrt{d}}\right)V,
\end{equation}
where $A$, $Q$, $K$, and $V$ represent the cumulative interactions,
queries, keys and values as matrices, and $d$ is the dimensionality of
the key vectors.  As in the original paper, we use 4 independent such
self-attention heads.  Subsequently, the output vectors per head are
concatenated and passed on to a fully-connected layer, and finally
added to the entity vector $e_i$ via residual connection to yield a
fixed-size embedding vector per entity.

The Transformer operation thus produces embeddings which encode
relations between loot in the environment. In this case we represent
each object by an array of its attribute bonuses, normalized between 0
and 1, which are further processed by fully connected layers with
shared weights across loot types. Based on these representations, a
Transformer layer is applied to reason about loot-loot relations,
resulting in a fixed-size embedding per entity. Following the concept
of \textit{spatial encoders} from AlphaStar, all entity
representations are then scattered into a spatial map so that the
embedding at a specific location corresponds to the unit/object placed
there.

More specifically, we create an empty map and place the embeddings
returned by the Transformer at the corresponding positions where the
loot is located in the game. We produce such scattered maps for both
global and local views which, as before, are concatenated with the
embedding map of categorical tile type and then passed on to the
remaining convolutional layers. The full network is shown in
figure \ref{newnets}b.  This model is, again, independent from changes
to the loot generation system, and even if developers change the
number of attributes during production, this architecture does not
require any adaptation, but can simply be retrained on the new
game. The biggest weakness is that this architecture is quite complex
and requires more computational resources, which goes against the last
desideratum defined in section \ref{sec:models}.

\section{Experimental results}
\label{sec:experiments}
In this section we report on experiments performed to evaluate
differences, advantages, and disadvantages of the two new
architectures with respect of the Categorical network. All of our
policy networks were implemented using the Tensorforce
library~\cite{kuhnle17}.\!\footnote{Code available
at \url{http://tiny.cc/ad_npc}}

We follow the same training setup of our original DeepCrawl work. At
the beginning of each episode, the shape and the orientation of the
map, as well as the number of impassable and collectible objects and
their positions are randomly generated; the initial position of the
player and the agent is random, as well as their initial equipment. We
also use curriculum learning \cite{bengio09} with the same phases as
the original paper and during training the agents fight against an
opponent that always makes \textit{random moves}. The only difference
is the addition of the \textit{loot generation system} described in
section~\ref{sec:loot-system}: each collectible item now has four
attributes which correspond to and modify the actor properties (HP,
DEX, ATK, DEF). At the beginning of each episode these values are
drawn randomly from a uniform distribution for each loot object on the
map.

We trained three NPC classes (Archer, Warrior, and Ranger (the same as
those from the original DeepCrawl paper) using the Transformer, Dense
Embedding, and the original Categorical deep policy networks. The NPC
classes are distinguished from one another by their character
attributes -- see \cite{sesto19} for a complete description of the
training procedure. In the following, we assess each of the main
requirements discussed above in section~\ref{sec:desiderata} in light
of our experimental results.

\minisection{Performance.} Figure \ref{training} shows the training
curves for our two proposed policy networks. The two architectures
achieve the same reward, demonstrating that both are able to properly
handle the new version of the environment. Table \ref{allvsall} shows
that, if two agents of the same class fight each other in the testing
configuration (where they start with the max amount of HP and their
initial equipment are neutral weapons, while the loot in the map is
still procedural), the Transformer based policy has a slight advantage
against the Dense Embedding network.

\renewcommand{\arraystretch}{1.3}
\begin{table*}[ht!]
  \caption{ Success rates averaged over 100 episodes for pairs of
    policy networks playing against each other in different testing
    environments.  \textit{Procedural Loot} refers to the environment
    with fully procedural loot, where each item attribute is drawn
    randomly at the beginning of an episode.  \textit{Uniform loot}
    refers to the variant with a fixed set of only three types of
    weapons (low, medium and high power ones). \textit{Skewed Loot}
    refers to another another such variant, one in which the strongest
    weapons are far more powerful than the other two. For more details
    about the experimental setup and loot distributions, see
    section \ref{sec:generalization}. The proposed network
    architectures generalize better across distributional loot changes
    compared to the original categorical architecture.
    }
  \label{allvsall}
  \begin{center}
    \begin{small}
      
        \scalebox{0.85}{
          \begin{tabular}{c|c|c|c||c|c|c||c|c|c}
            \toprule
            & \multicolumn{3}{c||}{\thead{Transformer vs Dense Embedding}} 
            & \multicolumn{3}{c||}{\thead{Dense Embedding vs Categorical}}
            & \multicolumn{3}{c}{\thead{Transformer vs Categorical}} \\
            
            & \thead{Procedural\\loot} & \thead{Uniform\\ loot} & \thead{Skewed\\ loot} 
            & \thead{Procedural\\loot} & \thead{Uniform\\ loot} & \thead{Skewed\\ loot}
            & \thead{Procedural\\loot} & \thead{Uniform\\ loot} & \thead{Skewed\\ loot} \\ 
            
            \midrule
            
            Archer 
            & \makecell{55\%} & \makecell{56\%} & \makecell{52\%} 
            & \makecell{62\%} & \makecell{58\%} & \textbf{\makecell{67\%}} 
            & \makecell{60\%} & \makecell{57\%} & \makecell{66\%} \\

            Warrior 
            & \makecell{50\%} & \makecell{52\%} & \makecell{50\%} 
            & \makecell{60\%} & \makecell{56\%} & \textbf{\makecell{66\%}} 
            & \makecell{60\%} & \makecell{58\%} & \makecell{66\%} \\
            
            Ranger 
            & \makecell{52\%} & \makecell{50\%} & \makecell{49\%}
            & \makecell{58\%} & \makecell{56\%} & \makecell{63\%} 
            & \makecell{56\%} & \makecell{58\%} & \textbf{\makecell{64\%}} \\
            
            \bottomrule
          \end{tabular}}
      
    \end{small}
  \end{center}
\end{table*}
\renewcommand{\arraystretch}{1.0}

We cannot compare these training curves with those in the original
work because of the dynamic nature of the environment introduced by
the procedural loot system. The only way to compare with the
Categorical network is to train agents with it in the new environment
after discretizing the loot attributes into potentially very many
unique object IDs. We can, however, have agents of the same class but
with different policy fight each other in this new environment. As
table
\ref{allvsall} shows, the proposed policies have higher average
success rate with respect to the Categorical policy network. This
demonstrates that these solutions better capture the differences
between loot objects.

\minisection{Adaptation.}
\label{sec:generalization}
To demonstrate the improved generalization capacity of our proposed
network architectures, we tested them by changing the loot
distribution from the fully procedural one used during training to a 
fixed distribution. This new environment has only
three different type of weapons: low, medium and high power (both
ranged and melee) that have clear differences between each other --
similar to the fixed weapons in the original
DeepCrawl. For \textit{high power} weapons we mean loot that gives
high value bonuses for all attributes, and so forth for medium and low
power ones.  Based on the four attributes in DeepCrawl, in this
variant a high power sword has attribute bonuses of \texttt{[+2, +2,
+2, +2]}, a medium power sword has
\texttt{[+0, +0, +0, +0]}, and a low power sword 
\texttt{[-2, -2, -2, -2]}.
We refer to this distribution 
as the \textit{uniform loot distribution}. We then compare the agents,
which have been trained with the full procedural loot, in this testing
environment. As table \ref{allvsall} shows, our proposed models have a
small advantage compared to the original Categorical framework.

In a subsequent experiment, we change the distribution of fixed loot
power: there are still three weapon types, low, medium and high power,
but the high power weapons are \emph{far} more powerful than the
medium and low power ones, which are comparatively similar. More
concretely, a high power sword here has attribute bonuses of
\texttt{[+5, +5, +5, +5]}, a medium sword \texttt{[-2, -2, -2, -2]}
and a low power sword \texttt{[-3, -3, -3, -3]}.
We call this distribution 
the \textit{skewed loot distribution}. As shown in table \ref{allvsall}, with this
configuration the success rate for our proposed architectures is much higher,
outperforming agents trained with the previous framework and hence showing
better adaptation than the Categorical architecture to such a change in
the balance of the game.

\minisection{Scalability.} To handle the procedural loot system with the
Categorical architecture developers must define a fixed set of class
IDs prior to training. This is not a trivial task and can quickly
become intractable when the number of attributes for each object
increases. Moreover, since the Categorical framework does not
generalize (see table~\ref{allvsall}), if developers want to change
loot generation they must define a new set of classes, and that forces
retraining of agents. Instead, with our proposed solutions developers
simply train their agents with procedurally generated loot and can
decide after training whether to use, in the final game, random
objects or a fixed set of weapons for balancing the game: our agents
will manage both situations without the need of retraining.

Both the proposed frameworks properly handle changes in the number of
attributes. In this case retraining is mandatory, but developers need
not to worry about changing the network architectures: with the Dense
Embedding network they need only to add a new channel for each new
attribute, while the Transformer based does not require any changes
since it is completely independent of loot parameterization.  The
Transformer embedding can even handle loot with various number of
attributes per type, providing a big advantage with respect to Dense
Embedding which requires loot with the same number of attributes.

The biggest drawback of the Transformer network is its complexity.
While the Dense and Categorical embedding networks require about 1.3
minutes to train 100 episodes, the Transformer network takes twice as
long. The average training times for 100 episodes are 3.31 minutes,
1.36 minutes and 1.10 minutes for the Transformer, Dense Embedding and
Categorical networks, respectively.  Training was performed on an
NVIDIA RTX 2080 SUPER GPU with 8GB of RAM.

In a video game design and development context, this is an important
aspect to consider: the continuous changes in the gameplay mechanics
require many retrainings, and having a small network is essential. In
addition, these frameworks must be implemented to target devices with
reduced performance, increasing the need for small and efficient
models.

\subsection{Optimization and hyperparameters}
We use the Proximal Policy Optimization algorithm~\cite{schulman17} to
optimize the agent model. The agent is trained over the course of
multiple episodes, each of which lasts at most 100 steps. An episode
may end with the agent achieving success (i.e. agent victory), failure
(i.e. agent death) or reaching the maximum step limit. After every
fifth episode, an update of the agent weights is performed based on
the previous five episodes.  PPO is an Actor-Critic algorithm with two
networks to be learned: the actor policy and the critic state-value
function. For both the dense embedding and transformer variant, the
critic uses the same structure as the policy network.

Hyperparameter values are the same in all experiments and were chosen
after a preliminary set of experiments with different configurations:
the policy learning rate $lr_p = 5 \cdot 10^{-6}$, the baseline
learning rate $lr_b = 5 \cdot 10^{-4}$, the agent exploration rate
$\epsilon = 0.2$, and the discount factor $\gamma = 0.99$. 
For the transformer architecture, we used a two-headed
self-attention
layer, with queries, keys and values of size $32$ and a two-layers MLP
with size $128$ and $32$.

\section{Conclusions}
\label{sec:conclusion}

In this paper, we described several extensions to our DeepCrawl framework
\cite{sesto19}. First, we implemented a procedural loot generation system which
augments the game with a degree of complexity that makes the game more
compelling as benchmark for DRL algorithms, particularly in the
context of game development. Moreover, we proposed two neural network
architectures, one based on Dense Embeddings and one based on
Transformers, which both show substantially improved performance due
to their capabilities to reason about loot and attribute
bonuses. Overall, our experimental analysis slightly favors the Dense
Embedding approach due to its reduced complexity and computational
requirements.

The advantages for game development are twofold. On the one hand,
Roguelikes such as DeepCrawl may contain a large number of items, or
indeed employ a procedural loot generation system, so the ability to
effectively learn how to compare and prioritize loot is important for
NPCs. On the other hand, this ability makes NPCs robust to
modifications to the loot system during development, without the need
to retrain the behavioral models from scratch every time. This is
important, first, for the balancing process during playtesting which
is crucial to final quality; and second, both our proposed
architectures can easily be adapted in the face of major changes to
the loot system which may occur during production.

\bibliography{main}

\end{document}